\documentclass[letterpaper, 10 pt, conference]{ieeeconf}  

\IEEEoverridecommandlockouts                              

\usepackage{algorithm}
\usepackage[noend]{algpseudocode}
\usepackage{amsmath}
\usepackage{amssymb}
\usepackage{booktabs}
\usepackage{caption}
\usepackage{diagbox}
\usepackage{dblfloatfix}
\usepackage{graphicx} 
\usepackage{hhline}
\usepackage{mathtools}
\usepackage{multirow}
\usepackage{subcaption}
\usepackage[font=small]{caption}
\usepackage{tabularx}
\usepackage{xcolor}

\usepackage{tikz}

\newcommand\copyrighttext{%
  \footnotesize \textcopyright 2022 IEEE.  Personal use of this material is permitted.  Permission from IEEE must be obtained for all other uses, in any current or future media, including reprinting/republishing this material for advertising or promotional purposes, creating new collective works, for resale or redistribution to servers or lists, or reuse of any copyrighted component of this work in other works.}
\newcommand\copyrightnotice{%
\begin{tikzpicture}[remember picture,overlay]
\node[anchor=south,yshift=10pt] at (current page.south) {\fbox{\parbox{\dimexpr\textwidth-\fboxsep-\fboxrule\relax}{\copyrighttext}}};
\end{tikzpicture}%
}

\title{\LARGE \bf
Automating Reinforcement Learning with Example-based Resets
}

\author{Jigang Kim$^{1,2}$, J. hyeon Park$^{1,2}$, Daesol Cho$^{1,2}$ and H. Jin Kim$^{1,2}$
\thanks{$^{1}$Department of Mechanical and Aerospace Engineering, Seoul National University, Seoul, Korea}%
\thanks{$^{2}$Automation and Systems Research Institute (ASRI), Seoul, Korea}%
}

\begin{document}

\bstctlcite{BSTcontrol} 

\maketitle
\copyrightnotice 
\thispagestyle{empty}
\pagestyle{empty}

\newcommand{\reducemargin}{-0.5cm}

\begin{abstract}
Deep reinforcement learning has enabled robots to learn motor skills from environmental interactions with minimal to no prior knowledge. However, existing reinforcement learning algorithms assume an episodic setting, in which the agent resets to a fixed initial state distribution at the end of each episode, to successfully train the agents from repeated trials. Such reset mechanism, while trivial for simulated tasks, can be challenging to provide for real-world robotics tasks. Resets in robotic systems often require extensive human supervision and task-specific workarounds, which contradicts the goal of autonomous robot learning. In this paper, we propose an extension to conventional reinforcement learning towards greater autonomy by introducing an additional agent that learns to reset in a self-supervised manner. The reset agent preemptively triggers a reset to prevent manual resets and implicitly imposes a curriculum for the forward agent. We apply our method to learn from scratch on a suite of simulated and real-world continuous control tasks and demonstrate that the reset agent successfully learns to reduce manual resets whilst also allowing the forward policy to improve gradually over time.
\end{abstract}

\section{Introduction}
Deep reinforcement learning (RL) methods have shown much promise in learning complex skills in the absence of prior knowledge both in simulation \cite{iros2020:duan2016benchmarking} and in the real world \cite{iros2020:gu2017deep}. Recent advances in large-scale RL applications such as playing real-time strategy games \cite{iros2020:vinyals2019grandmaster}\cite{iros2020:berner2019dota} and dexterous manipulation of objects \cite{iros2020:andrychowicz2020learning} demonstrate the potential of RL methods. However, most RL algorithms are not specifically designed with learning in the real world in mind, making assumptions that present a challenge for autonomous robot learning. One major hurdle to autonomy is the often overlooked reset mechanism. Conventional RL algorithms assume the ability to sample from the initial state distribution which does not hold for real environments.

Previous attempts to apply RL to real robots have relied on some combination of human intervention, scripted actions, and task-specific experiment setups to implement a reset mechanism \cite{ral2021:yahya2017collective}\cite{ral2021:sharma2020emergent}\cite{ral2021:kalashnikov2018qt}. Some prior works even design mechanical rigs with actuators and sensors to minimize human interventions during resets \cite{iros2020:zeng2019tossingbot}. For tasks such as goal reaching or trajectory tracking, scripted actions may be enough to prevent most human interventions but still require human oversight to handle edge cases. Tasks involving manipulation may require additional environmental setup to prevent out-of-reach objects and human intervention to manually reconfigure objects. However, these tailored reset mechanisms based on prior knowledge of the task are workarounds to apply RL to real robots and do not address the incompatibility between RL algorithms and autonomous robot learning.

\begin{figure}[t]
    \centering
    \includegraphics[width = 0.89\linewidth]{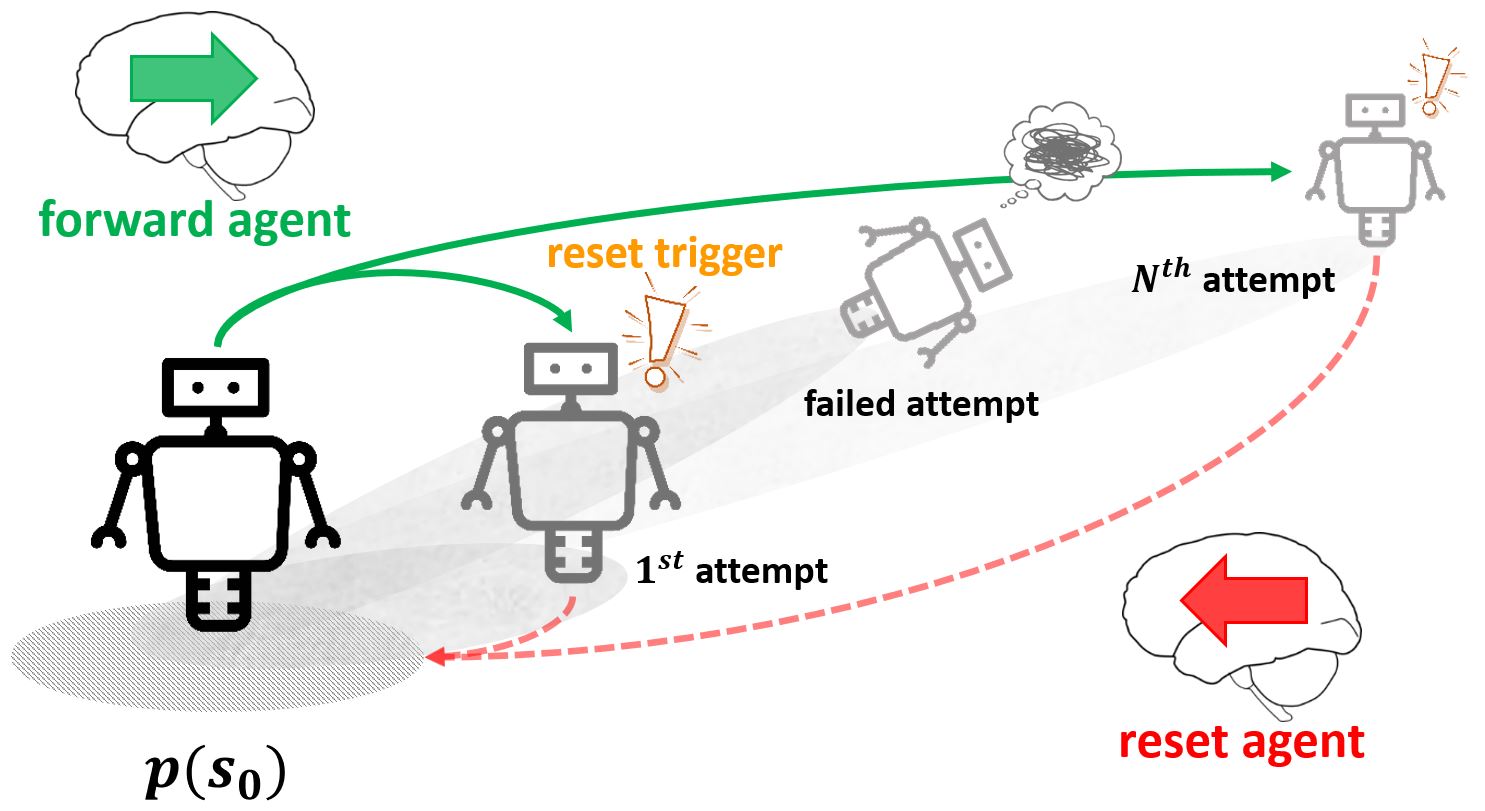}
    \caption{We consider a reset agent for autonomous and continuous training. Resets are triggered to prevent manual resets and impose an implicit curriculum on the forward agent.}
    \label{fig:thumbnail}
    \vspace{\reducemargin}
\end{figure}

We incorporate resets as part of the learning process to address the issue of autonomy and provide a natural extension to existing RL methods. In addition to the conventional RL agent (forward agent) that learns a given task, a reset agent can be trained to return to the initial state distribution (Fig. \ref{fig:thumbnail}). This enables continuous training on real robots by learning from both forward and reset episodes, as opposed to conventional RL where training is halted during resets. Furthermore, allowing the reset agent to preemptively trigger resets instead of waiting for the forward episode to terminate has additional benefits. Value-based reset trigger can prevent the forward policy from leading the system into states from which the reset policy cannot reset. It also implicitly generates a curriculum for the forward agent by confining it to the vicinity of the initial state in the early stages of training when the reset agent is not yet capable and gradually allowing the forward agent to explore further as the reset agent improves.

In this paper, we train the reset agent from examples of initial states instead of rewards. Designing a reset reward function depends on the given task, similar to how tailored reset mechanisms rely on prior knowledge. In contrast, examples of initial states can simply be collected at the beginning of every forward episode, which is a task-agnostic process. Also, learning to reset from examples allows for a better extension to conventional RL algorithms since the reset agent works under the hood without additional design effort by the end-user. We take advantage of recent advances in example-based control to train the reset agent.

The main contribution of our work is in extending the RL framework for robot learning towards greater autonomy by learning a reset agent in a self-supervised manner with examples of initial states. We apply our method to learn from scratch on a suite of simulated and real-world continuous control tasks to demonstrate that the reset agent successfully learns to reset from examples of initial states and triggers resets to reduce manual resets and prevent irrecoverable states whilst also allowing the forward policy to improve gradually over time. We highlight our contribution of real robot deployment of the proposed autonomous RL framework and claim that it is a step towards fully autonomous robot learning.

\section{Related Work}
As RL is increasingly applied to robotic applications, there has been a growing interest in automating the training process and minimizing the human factor \cite{iros2020:zeng2019tossingbot}\cite{ral2021:ha2021learning}. Recent literature from multi-task and goal-conditioned RL that reformulate resets \cite{ral2021:xu2020continual}\cite{ral2021:zhu2020ingredients}\cite{ral2021:gupta2021reset}\cite{ral2021:sharma2021persistent}\cite{ral2021:eysenbach2020rewriting} and works on multi-stage task methods that sequentially roll out forward and reset policies \cite{ral2021:han2015learning}\cite{iros2020:eysenbach2018leave}\cite{ral2021:smith2020avid} are closely related to our work. Works on value-based implicit curriculum \cite{ral2021:sharma2021persistent}\cite{ral2021:thananjeyan2021recovery}\cite{ral2021:schaul2016prioritized} and reversibility in RL \cite{ral2021:rahaman2020learning}\cite{ral2021:grinsztajn2021there} are also relevant to our work.

Concerns of autonomy in RL due to the assumption of a reset mechanism have been raised previously. Examples include learning diverse primitives \cite{ral2021:xu2020continual} or using a perturbation controller \cite{ral2021:zhu2020ingredients} in order to ``reset" to a broad set of initial states and reformulating resets into goal-conditioned RL by leveraging a sub-goal generator \cite{ral2021:sharma2021persistent}. These works assume a reversible environment to circumvent the need for a reset in the traditional sense---as in returning to some fixed initial state---allowing agents to learn from various states without the danger of getting stuck. In contrast, our method makes no such assumption on the environment by preventing the agent from irrecoverable states with reset triggers.

Value-based trigger has been featured in prior works involving multiple agents. For example, it has been applied to safe RL where a recovery policy is executed if the pre-trained safety critic deems the action of the forward policy to be unsafe \cite{ral2021:thananjeyan2021recovery}. Similarly, our method triggers resets based on the reset critic to prevent the forward agent from manual resets, but it is learned along with the forward agent and not pre-trained. Implicit curriculum imposed on the forward agent by the learned reset trigger resembles reverse curriculum \cite{iros2020:florensa2017reverse}\cite{iros2020:zakka2019form2fit}, but our method differs in allowing such behavior to emerge as a result of joint training of the forward and reset agents.

We build upon prior work \cite{iros2020:eysenbach2018leave} which required additional handcrafted reward function to specify the reset objective. Our method eliminates the need for such reset reward by adopting example-based control that learns from examples of initial states. Thus, unlike prior work which relied on environment-dependent reset Q-function threshold to trigger resets, example-based control methods enable learning of a reset critic that estimates the probability of returning to the initial state, an environment-agnostic metric, to trigger resets and train the policy.

\section{Preliminaries}
Typical RL setting involves a sequential decision making problem represented by a Markov decision process (MDP). MDP is defined by the tuple ($\mathcal{S}$, $\mathcal{A}$, $\mathcal{P}$, $r$, $p(s_{0})$), where $\mathcal{S}$ is the state space, $\mathcal{A}$ is the action space, $\mathcal{P}$ is the transition dynamics $P(s'|s,a)$, $r(s,a)$ is the scalar reward function, and $p(s_{0})$ is the initial state distribution. The objective of RL algorithms is to find the optimal policy $\pi^{*}(a|s)$ that maximizes the expected $\gamma$-discounted return $\mathbb{E}_{\pi}[\sum_{t=0}^{\infty} \gamma^{t}r(s_{t},a_{t})]$.

\subsection{Event framework for the reset agent}
We adopt the event framework \cite{iros2020:fu2018variational}\cite{ral2021:eysenbach2021replacing} that casts control as a probabilistic inference problem to learn to reset from examples of initial states $\mathcal{S}^{*}=\{s^{*} \sim p(s_{0})\}$ without external rewards. The event variable defined by the binary random variable $e_t \in \{0,1\}$ indicates whether the desired outcome of reaching the initial state has been achieved at time $t$ and $p(e_{t}=1|s_{t})$ is the probability of achieving the desired outcome. The objective is to find the optimal policy that maximizes the probability of achieving the desired outcome in the future $p^{\pi}(e_{t:\infty}=1)$. The problem setting of the event framework resembles that of goal-conditioned RL with sparse reward. However, unlike goal-conditioned RL which relies on goal-relabeling to diverse goals (hindsight), the event framework directly recovers a dense reward from a small set of desired outcomes and is rooted in probability. Thus, it is better suited for the reset agent that learns to reach a small subset of the state space (initial state) and allows for a more intuitive reset trigger based on the probability of returning to the initial state. We adopt RCE \cite{ral2021:eysenbach2021replacing}, which directly learns the value function from desired outcomes and transitions without the explicit use of learned rewards by deriving an update rule based on first principles.

\subsection{Recursive Classification of Examples (RCE)}
RCE defines a future success classifier $C^{\pi}_{\theta}(s_{t},a_{t})$ that discriminates between state-action pairs from the conditional distribution $p^{\pi}(s_{t},a_{t}|e_{t:\infty}=1)$ and the marginal distribution $p(s_{t},a_{t})$ with respective weights $p(e_{t:\infty}=1)$ and 1,
\begin{align*}
    C^{\pi}_{\theta}(s_{t},a_{t}):=\dfrac{p^{\pi}(s_{t},a_{t}|e_{t:\infty}=1)p(e_{t:\infty}=1)}{p^{\pi}(s_{t},a_{t}|e_{t:\infty}=1)p(e_{t:\infty}=1) + p(s_{t},a_{t})}
\end{align*}
such that the probability of achieving the desired outcome in the future given current state and action $p^{\pi}(e_{t:\infty}=1|s_{t},a_{t})$ can be recovered with a simple classifier ratio:
\begin{align}
    p^{\pi}(e_{t:\infty}=1|s_{t},a_{t})=\dfrac{C^{\pi}_{\theta}(s_{t},a_{t})}{1-C^{\pi}_{\theta}(s_{t},a_{t})}
\label{eqn:classifier_ratio}
\end{align}

The loss for the future success classifier involves two cross-entropy loss terms, the first term assigning a label of 1 for desired outcomes and the second term assigning a label of $\frac{\gamma\omega^{(t)}}{\gamma\omega^{(t)}+1}$ for transitions from the replay buffer,
\begin{align}
\label{eqn:classifier_loss}
    &\min_{\theta} (1-\gamma)\mathbb{E}_{\substack{p(s_{t}|e_{t}=1), \\ a_{t} \sim \pi(a_{t}|s_{t})}}[\mathcal{CE}(C^{\pi}_{\theta}(s_{t},a_{t});1)] \\ &+ (1+\gamma\omega^{(t)})\mathbb{E}_{p(s_{t},a_{t},s_{t+1})}\left[\mathcal{CE}\left(C^{\pi}_{\theta}(s_{t},a_{t});\dfrac{\gamma\omega^{(t)}}{\gamma\omega^{(t)}+1}\right)\right] \nonumber
\end{align}
where $\omega^{(t)}$ is the classifier ratio at the next time step,
\begin{align}
    \omega^{(t)} = \mathbb{E}_{p(a_{t+1}|s_{t+1})}\left[\dfrac{C^{\pi}_{\theta}(s_{t+1},a_{t+1})}{1-C^{\pi}_{\theta}(s_{t+1},a_{t+1})}\right]
\label{eqn:classifier_omega}
\end{align}
This replaces the MSE TD loss typical of critic update in standard actor-critic RL algorithms. We refer the reader to Section 3.2 of \cite{ral2021:eysenbach2021replacing} for derivation details. Actor update is performed by gradient ascent on the policy network using action gradients with respect to the future success classifier.

\section{Methods} \label{methods}
Our method simultaneously learns a forward agent and a reset agent. The forward agent consists of a policy $\pi_{f}(s)$ and a Q-function $Q^{\pi_{f}}(s,a)$. Likewise, the reset agent includes $\pi_{r}(s)$ but instead of a Q-function includes a reset success classifier $C^{\pi_{r}}(s,a)$ which is a proxy for the probability of returning to the initial state $p^{\pi_{r}}(e_{t:\infty}=1|s_{t},a_{t})$ defined by the classifier ratio (Equation \ref{eqn:classifier_ratio}). The forward agent performs a task defined by the reward signal from the environment following the existing RL setting. The reset agent, which acts as a safety layer between the forward policy and the environment, directly learns the reset success classifier without rewards (Equation \ref{eqn:classifier_loss}). We adopt DDPG \cite{iros2020:lillicrap2015continuous} for both agents but other off-policy actor-critic variants are also applicable.

During training, forward and reset agents run in an alternating fashion. The forward agent rolls out the forward policy until it reaches the maximum step limit and requests a reset or until the reset agent preemptively triggers a reset. For every forward policy action $a_{t}$, the reset agent evaluates $p^{\pi_{r}}(e_{t:\infty}=1|s_{t},a_{t})$ and triggers a reset if its value is less than some threshold $p_{thresh}$. When a reset is either requested or triggered, the reset agent then runs the reset policy instead of immediately resetting to the initial state with a manual reset. The reset policy is rolled out until the system reaches the initial state distribution. If it fails to do so within the maximum step limit, a manual reset is inevitable. In either case, the reset agent ends up at the initial state and can collect an example of the initial state at the beginning of every forward episode. The collected examples along with transitions from $\mathcal{B}_{r}$ are used to train the reset agent in a self-supervised manner (Equation \ref{eqn:classifier_loss}). This is a general overview of the training scheme and further details are provided in Algorithm \ref{algorithm:overview}.

Note that under this training scheme the forward agent is not directly penalized for triggering resets. However, if the forward policy triggers a reset without achieving the task, the corresponding value for the policy would remain low and the forward policy, which is continually updated to increase the value, will eventually learn to achieve the task without triggering resets. This makes sense, especially if we consider the mechanics of the reset agent to be part of the environment with which the forward agent interacts. Also, this setting allows for a seamless extension to conventional RL as it does not modify the forward agent.

\begin{algorithm}[t]
\caption{Automating RL with example-based resets}\label{algorithm:overview}
\begin{algorithmic}[1]
\State \textbf{Initialize} networks $\pi_{f}(s)$, $Q^{\pi_{f}}(s,a)$, $\pi_{r}(s)$
\State \textbf{Initialize} network ensemble $C_{i}^{\pi_{r}}(s,a)$ ($i=1,\dots,K$)
\State \textbf{Initialize} initial state buffer $\mathcal{S}^{*}$, replay buffers $\mathcal{B}_{f}$, $\mathcal{B}_{r}$
\State \textbf{Initialize} environment with $env.reset()$
\While{not total steps reached}
\State save an example of initial state to $\mathcal{S}^{*}$
    \While{not reset} \Comment{forward episode}
        \State $a_{t}\gets \pi_{f}(s_{t})$
        \State evaluate $p^{\pi_{r}}_{i}(e_{t:\infty}=1|s_{t},a_{t})$ from $C_{i}^{\pi_{r}}(s_{t},a_{t})$
        \If{$\overline{p}^{\pi_{r}}(e_{t:\infty}=1|s_{t},a_{t})<p_{thresh}$}
            \State trigger reset
        \ElsIf{reached max forward episode steps}
            \State request reset
        \Else
            \State $(s_{t+1}, r_{t})\gets env.step(a_{t})$
            \State save transition to $\mathcal{B}_{f}$ and update $\pi_{f}$, $Q^{\pi_{f}}$ 
        \EndIf
    \EndWhile
    \While{not at initial state} \Comment{reset episode}
        \State $a_{t}\gets \pi_{r}(s_{t})$
        \If{under max reset episode steps}
            \State $s_{t+1}\gets env.step(a_{t})$
            \State save transition to $\mathcal{B}_{r}$ and update $\pi_{r}$, $C_{i}^{\pi_{r}}$
        \Else \Comment{initial state not reached until max step}
            \State manual reset with $env.reset()$
        \EndIf
    \EndWhile
\EndWhile
\end{algorithmic}
\end{algorithm}

We make the following assumptions similar to \cite{iros2020:eysenbach2018leave}. The initial state distribution is assumed to be unimodal to prevent multiple objectives for the reset agent and it is assumed to be possible to accomplish the forward task without falling into irrecoverable states. We additionally assume that the reset agent is not given feedback (i.e. rewards) from the environment on how to reset but is notified when it has successfully reset. Note that classifying initial states is much easier than assigning a proper credit on how to reset.

\subsection{Reset trigger mechanism}
Our goal is to reduce manual resets with the reset agent. The reset agent should trigger a reset by evaluating whether the action of the forward policy will lead to irrecoverable states. Note that the reset trigger should not only prevent states from which it cannot physically return (e.g. falling into a ditch, spilling the contents of a cup) but also states from which the reset policy is not yet capable of returning. For example, resets should be triggered almost immediately during the early stages of training when the reset policy is not yet capable. As the reset agent gathers examples of the initial state at the end of every reset episode, it should gain confidence and trigger fewer resets allowing the forward policy to explore even further.

Triggering resets by thresholding the output of a single network can lead to overestimation in not yet trained regions and fail to trigger resets. Thus, we use network ensembles $C^{\pi_{r}}_{i}(s_{t},a_{t})$ for the reset success classifier in order to estimate the epistemic uncertainty and trigger resets by evaluating the ensemble average of the probability of successful reset $\overline{p}^{\pi_{r}}(e_{t:\infty}=1|s_{t},a_{t})$. This combination encourages the reset agent to trigger resets under uncertainty from a lack of data, which is especially the case during the early stages of training. Naive ensembles are not varied enough so we incorporate a simple yet effective technique known as randomized prior functions (RPFs) \cite{iros2020:osband2018randomized} which represents the function approximator as a sum of two networks of the same structure where one learns the posterior (trainable part) and the other acts as the prior (untrainable part).

We provide justification for using the probability of successful resets to trigger resets by making a connection between $p^{\pi_{r}}(e_{t:\infty}=1|s_{t},a_{t})$ and the expected time steps $T$ to reach the initial state distribution. The discounted probability of successful reset at some future time given the reset policy $\pi_{r}$ is defined as,
\begin{align*}
    p^{\pi_{r}}(e_{t:\infty}=1|s_{t},a_{t})=\mathbb{E}_{p^{\pi_{r}}(s|s_{t},a_{t})}[p(e=1|s)]
\end{align*}
where $p^{\pi_{r}}(s|s_{t},a_{t})$ is the discounted future state distribution,
\begin{align*}
    p^{\pi_{r}}(s|s_{t},a_{t}):=(1-\gamma)\sum_{\Delta=0}^{\infty}\gamma^{\Delta}p^{\pi_{r}}(s_{t+\Delta}=s|s_{t},a_{t}) \nonumber
\end{align*}
Consider the following where states can be perfectly partitioned into initial states and non-initial states with corresponding $p(e=1|s)$ values of 1 and 0, respectively. If the reset policy $\pi_{r}$ reaches the initial state after $T$ steps and remains there indefinitely, then $p^{\pi_{r}}(e_{t:\infty}=1|s_{t},a_{t})$ becomes $\gamma^{T}$:
\begin{align*}
    p^{\pi_{r}}(e_{t:\infty}=1|s_{t},a_{t}) &= (1-\gamma)\sum_{\Delta=0}^{\infty}\gamma^{\Delta}p^{\pi_{r}}(e_{t+\Delta}=1|s_{t+\Delta}) \nonumber \\
    &= (1-\gamma)\left[\sum_{\Delta=0}^{T-1}0+\sum_{\Delta=T}^{\infty}\gamma^{\Delta}\right] = \gamma^{T} \nonumber
\end{align*}

\subsection{Implementation details}

Reset success classifier network ensemble presents a choice for updating the critic and the actor. The straightforward approach is to independently update each critic network $C^{\pi_{r}}_{i}$ with its bootstrapped label and update the actor network $\pi_{r}$ by evaluating the action gradient with respect to the ensemble average. Instead, we take the ensemble minimum for both critic and actor update to tackle the issue of overestimation, which not only is important for training the reset policy but also critical for the reset trigger mentioned previously. It has been shown that using n-step returns for the critic update improves the stability and performance of RCE. Thus, we also adopt n-step returns ($n=10$) and replace the label for the replay buffer transitions with the following:
\begin{align*}
    y = \dfrac{1}{2} \left( \dfrac{\gamma\omega^{(t)}}{\gamma\omega^{(t)}+1} + \dfrac{\gamma^{n}\omega^{(t+n-1)}}{\gamma^{n}\omega^{(t+n-1)}+1} \right)
\end{align*}
Note that setting $n=1$ recovers the original label from Equation \ref{eqn:classifier_loss}.
The reset success classifier $C^{\pi_{r}}$ should be in range $[0,0.5]$ due to the classifier ratio relation (Equation \ref{eqn:classifier_ratio}) and the fact that probabilities are in range $[0,1]$. We use the sigmoid parameterization for $C^{\pi_{r}}$ and clip the values when necessary, such as deriving the label for the RCE loss (Equation \ref{eqn:classifier_loss}) or evaluating the probability of successful reset for reset trigger.

\newcommand{\mainresultsubpath}{eval_proposed_shaded_blueblack}

\begin{figure*}[t]
    \vspace{\reducemargin}
    \centering
    \begin{subfigure}[b]{0.329\linewidth}
        \centering
        \includegraphics[width = 1\linewidth]{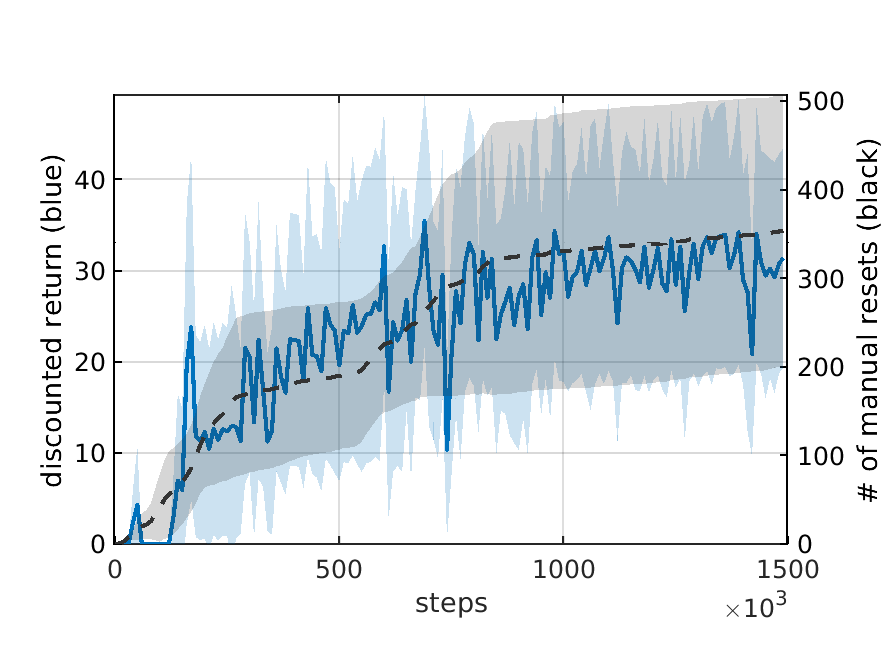}
        \caption{\textbf{\textit{ball-in-cup (catch)}}}
        \label{fig:main_result(ball-in-cup_catch)}
    \end{subfigure}
    \hfill
    \begin{subfigure}[b]{0.329\linewidth}
        \centering
        \includegraphics[width = 1\linewidth]{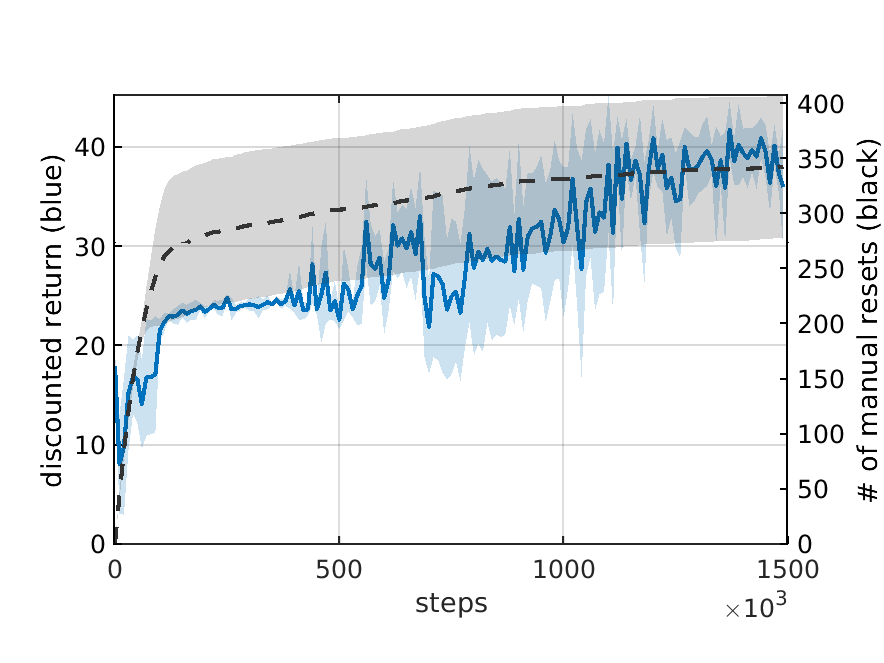}
        \caption{\textbf{\textit{peg-insertion (insert)}}}
        \label{fig:main_result(peg-insertion_insert)}
    \end{subfigure}
    \hfill
    \begin{subfigure}[b]{0.329\linewidth}
        \centering
        \includegraphics[width = 1\linewidth]{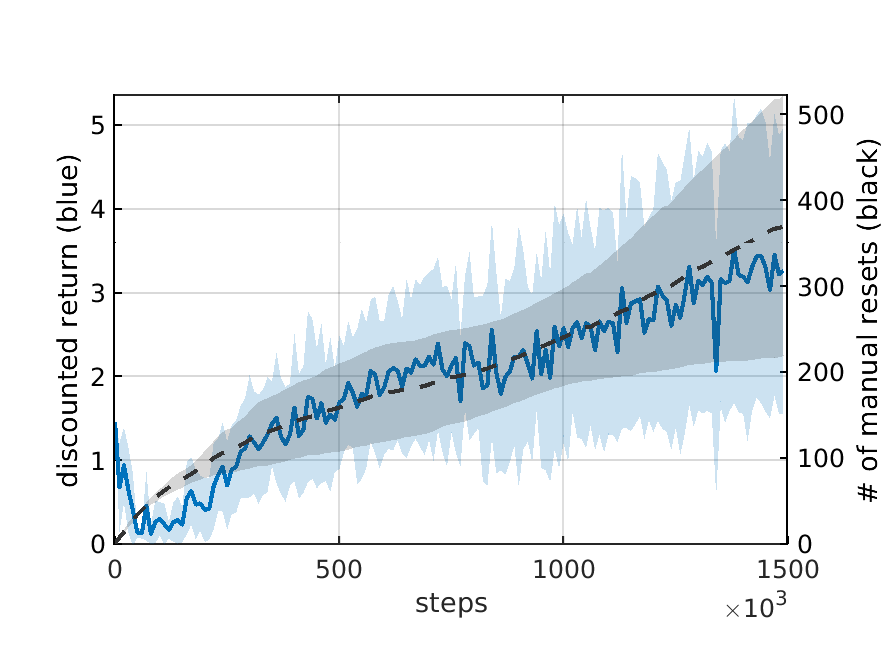}
        \caption{\textbf{\textit{cliff-cheetah (move forward)}}}
        \label{fig:main_result(cliff-cheetah)}
    \end{subfigure}
    \begin{subfigure}[b]{0.329\linewidth}
        \centering
        \includegraphics[width = 1\linewidth]{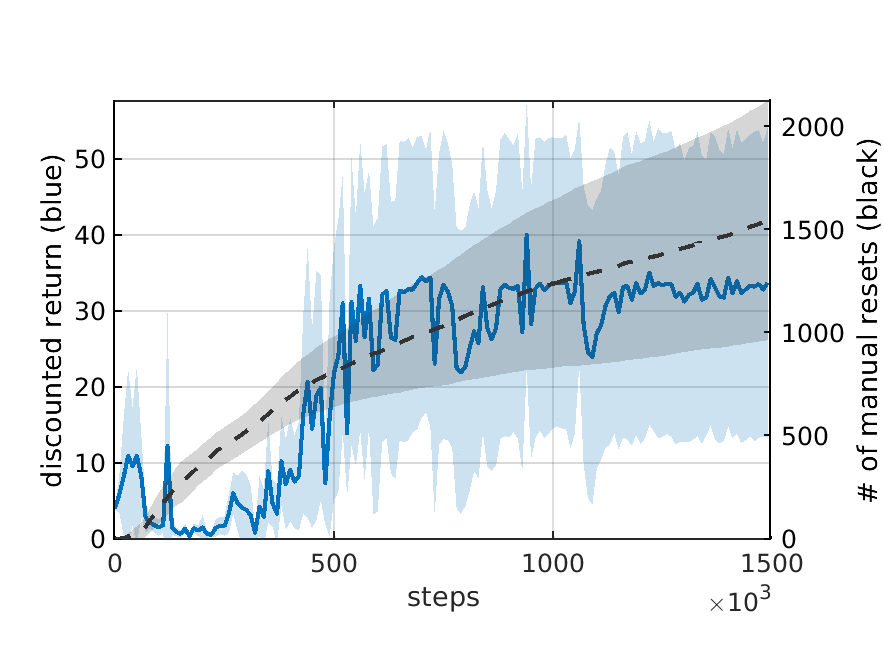}
        \caption{\textbf{\textit{ball-in-cup (throw)}}}
        \label{fig:main_result(ball-in-cup_throw)}
    \end{subfigure}
    \hfill
    \begin{subfigure}[b]{0.329\linewidth}
        \centering
        \includegraphics[width = 1\linewidth]{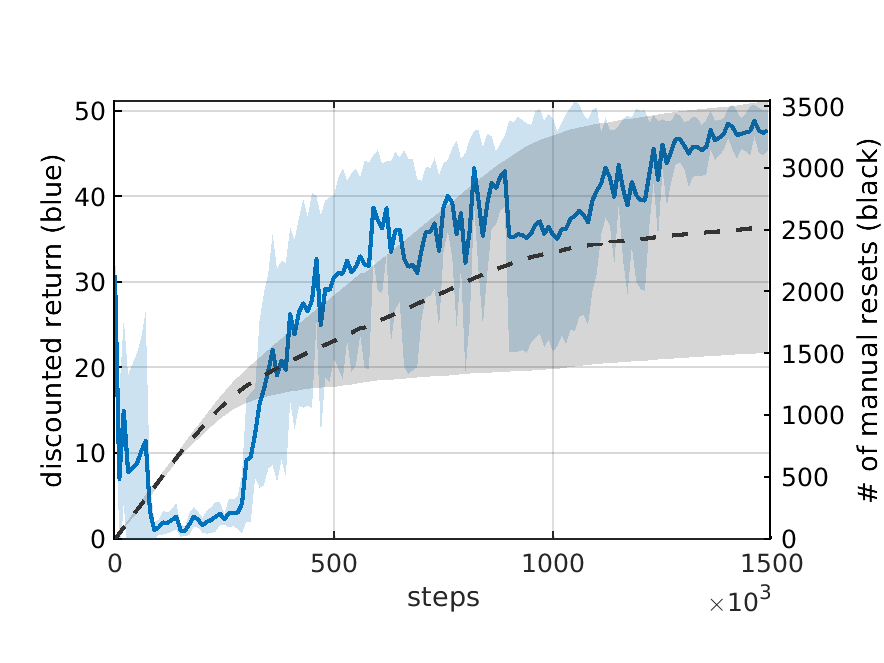}
        \caption{\textbf{\textit{peg-insertion (remove)}}}
        \label{fig:main_result(peg-insertion_remove)}
    \end{subfigure}
    \hfill
    \begin{subfigure}[b]{0.329\linewidth}
        \centering
        \includegraphics[width = 1\linewidth]{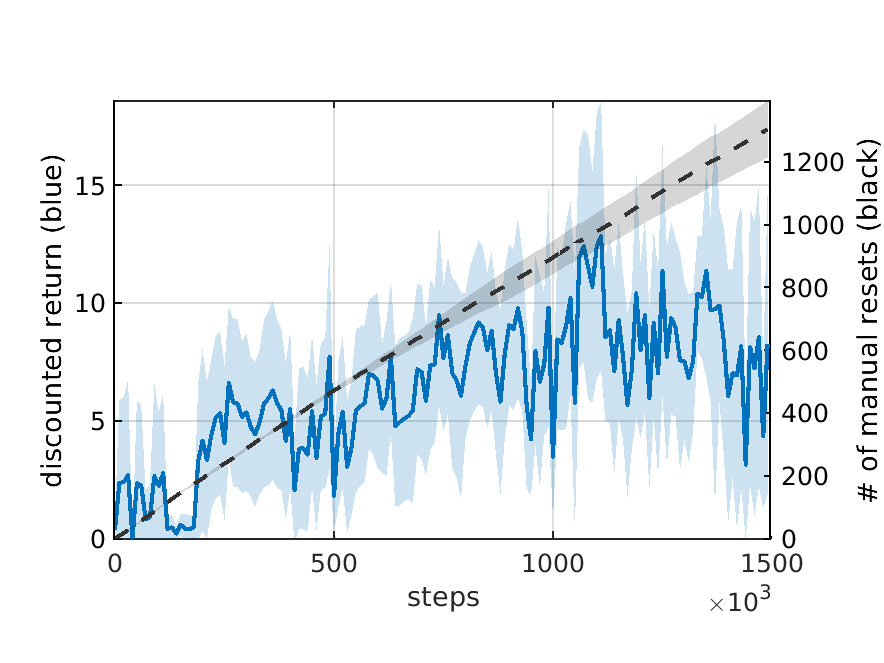}
        \caption{\textbf{\textit{cliff-walker (move forward)}}}
        \label{fig:main_result(cliff-walker)}
    \end{subfigure}
    \caption{Forward episodic return (blue solid line) and the number of manual resets (black dashed line) for six simulated tasks.}
    \label{fig:main_result}
    \vspace{\reducemargin}
\end{figure*}

\section{Experiments}\label{experiments}

We present empirical evidence to demonstrate that the reset agent reduces manual resets and learns to successfully reset with examples of initial states. To do so, we apply our method to various simulated continuous control tasks and compare it against prior work across several performance metrics. We also apply our method to learn from scratch on a real robot (robotic manipulator) to demonstrate the potential of the method. In the following sections, we provide hyperparameter details, describe the simulated and real-world tasks used for evaluation, and analyze the results.

\subsection{Hyperparameters}
The following settings were set as the default for all experiments unless otherwise specified. $p_{thresh}$ for the reset trigger was set to 0.1 and the reset agent was allowed up to double the number of maximum steps allowed for the forward agent before resorting to a manual reset. All networks were configured with two fully-connected hidden layers of dimensions 400 and 300 with ReLU activations. Policy networks additionally have $tanh$ activations on the output layer to bound the action. For Q-function and reset success classifier networks, only the states are fed into the input layer and the actions are later concatenated into the first hidden layer. Network ensembles were used for the reset agent with an ensemble of $K=5$ networks and an RPF scale factor of 3 for $C^{\pi_{r}}_{i}$. ADAM optimizer with a learning rate of 1e-4 was used for $\pi_{f}$ and $\pi_{r}$, and a learning rate of 1e-3 was used for $Q^{\pi_{f}}$ and $C^{\pi_{r}}_{i}$. Other hyperparameters include $\tau=$1e-3 for updating the target networks and a buffer size of 500k for both forward and reset agents.

\subsection{Simulated tasks}
The simulated environments are based on OpenAI Gym \cite{iros2020:openaigym2016} and DeepMind Control Suite \cite{iros2020:dmcontrol2018} environments. We evaluated on four environments and a variety of manipulation and locomotion tasks for a total of six tasks. \textbf{\textit{ball-in-cup}} is a standard environment from the Control Suite and its forward task was either to catch the ball (catch task) or to throw the ball as far as possible (throw task). The other three environments are based on OpenAI Gym. \textbf{\textit{peg-insertion}} involves a 7DoF manipulator holding a peg and its forward task was either to insert the peg into the hole (insert task) or to remove the peg from the hole (remove task). \textbf{\textit{cliff-cheetah}} and \textbf{\textit{cliff-walker}} are modified versions of the Gym MuJoCo environments HalfCheetah and Walker2d with the addition of a steep cliff at $x=14$ and $x=6$, respectively. The forward task of the cliff environments was to maintain some positive velocity. The dense forward task reward functions were normalized to $[0,1]$ which, combined with the discount factor $\gamma$ of 0.99, bounds the forward return to $[0,100)$. The maximum steps per episode were set to 100, except for the cliff environments which were set to 500 and their reset trigger threshold was adjusted accordingly to $p_{thresh}=0.05$.

\subsection{Simulation results}

We take a snapshot of the forward and reset agents at regular intervals during training, and roll out a pair of forward and backward episodes in a separate evaluation loop to evaluate the forward episodic return. Note that the action noise typical of DDPG to aid exploration was turned off for the evaluation loop. Performance metrics other than the forward episodic return were derived from the training loop. All of the simulation results were averaged across five random seeds.

\subsubsection{Overview of our method}

We applied our method to six simulated tasks and plot the forward episodic return and the number of manual resets, or in other words, the number of failed reset attempts by the reset agent (Fig. \ref{fig:main_result}). Multiple tasks are available for the \textbf{\textit{ball-in-cup}} and \textbf{\textit{peg-insertion}} environments, which allows us to observe the effect of task difficulty on learning. For \textbf{\textit{ball-in-cup}}, the throwing task is easier than the catch task and for \textbf{\textit{peg-insertion}}, the remove task is easier than the insert task. This is somewhat reflected in the episodic return, as the forward agent achieves slightly higher returns for the easier tasks in \textbf{\textit{ball-in-cup}} and \textbf{\textit{peg-insertion}} environments. The task difficulty for the reset agent is opposite to that of the forward agent since the reset agent has to ``undo" the forward task. Case in point, the number of manual resets for the \textbf{\textit{peg-insertion}} remove task is 10 times higher than that of the insert task since the remove task requires the reset agent to learn how to insert the peg. Such trends are also evident in \textbf{\textit{ball-in-cup}} but to a lesser degree.

One notable trend is that environments without irrecoverable states saw the number of manual resets mostly plateau (Fig. \ref{fig:main_result(ball-in-cup_catch)},\ref{fig:main_result(peg-insertion_insert)},\ref{fig:main_result(ball-in-cup_throw)},\ref{fig:main_result(peg-insertion_remove)}), whereas environments with irrecoverable states did not (Fig. \ref{fig:main_result(cliff-cheetah)},\ref{fig:main_result(cliff-walker)}). Cliff environments have a cliff at the end of the map where the reset agent is physically unable to reset from. The assumption that the forward agent can achieve the forward task without falling into irrecoverable states technically holds for these environments since the agent can learn to stop just before the fall. However, the dynamics of the forward agent encouraging risky behavior (inching closer to the cliff) while the reset agent is trying to prevent irrecoverable states can be an unstable one. In practice, the forward agent sometimes slips past the reset trigger and falls off the cliff.

\subsubsection{Comparison against the oracle/baseline}

\begin{table*}[t!]
\caption{Comparison of our method against LNT (oracle) and LNT-sparse (baseline).}
\label{table:results}
\begin{center}
\begin{tabularx}{1\linewidth}{c|*{12}{|X}}

\toprule

\multirow{2}{*}{\diagbox{\textbf{Algo.}}{\textbf{Env.}}} & \multicolumn{4}{c|}{\textbf{ball-in-cup}} & \multicolumn{4}{c|}{\textbf{peg-insertion}} & \multicolumn{4}{c}{\textbf{cliff-cheetah}} \tabularnewline
& \multicolumn{4}{c|}{(forward task: catch ball)} & \multicolumn{4}{c|}{(forward task: insert peg)} & \multicolumn{4}{c}{(forward task: move forward)} \tabularnewline

\hline

\multirow{2}{*}{} & \multicolumn{1}{c|}{average} & \multicolumn{1}{c|}{manual} & \multicolumn{1}{c|}{forward} & \multicolumn{1}{c|}{success} & \multicolumn{1}{c|}{average} & \multicolumn{1}{c|}{manual} & \multicolumn{1}{c|}{forward} & \multicolumn{1}{c|}{success} & \multicolumn{1}{c|}{average} & \multicolumn{1}{c|}{manual} & \multicolumn{1}{c|}{forward} & \multicolumn{1}{c}{success} \tabularnewline
& \multicolumn{1}{c|}{return} & \multicolumn{1}{c|}{resets} & \multicolumn{1}{c|}{share} & \multicolumn{1}{c|}{rate} & \multicolumn{1}{c|}{return} & \multicolumn{1}{c|}{resets} & \multicolumn{1}{c|}{share} & \multicolumn{1}{c|}{rate} & \multicolumn{1}{c|}{return} & \multicolumn{1}{c|}{resets} & \multicolumn{1}{c|}{share} & \multicolumn{1}{c}{rate} \tabularnewline

\hline

\textbf{LNT} & \centering 10.38 & \centering \textbf{276} & \centering 85.7\% & \centering \textbf{99.3\%} & \centering \textbf{30.50} & \centering \textbf{251} & \centering \textbf{71.7\%} & \centering \textbf{98.5\%} & \centering \textbf{4.77} & \centering 835 & \centering 15.0\% & \centering 85.7\% \tabularnewline
\textbf{ours} & \centering \textbf{19.51} & \centering 331 & \centering 79.7\% & \centering 96.8\% & \centering 29.20 & \centering 343 & \centering 66.3\% & \centering 96.6\% &  \centering 1.96 & \centering \textbf{371} & \centering \textbf{31.4\%} & \centering \textbf{94.8\%} \tabularnewline
\textbf{LNT-sparse} & \centering 5.48 & \centering 358 & \centering \textbf{89.0\%} & \centering 97.5\% & \centering 1.60 & \centering 7175 & \centering 2.6\% & \centering 74.4\% & \centering 2.96 & \centering 765 & \centering 15.4\% & \centering 80.3\% \tabularnewline

\midrule

\multirow{2}{*}{\diagbox{\textbf{Algo.}}{\textbf{Env.}}} & \multicolumn{4}{c|}{\textbf{ball-in-cup}} & \multicolumn{4}{c|}{\textbf{peg-insertion}} & \multicolumn{4}{c}{\textbf{cliff-walker}} \tabularnewline
& \multicolumn{4}{c|}{(forward task: throw ball)} & \multicolumn{4}{c|}{(forward task: remove peg)} & \multicolumn{4}{c}{(forward task: move forward)} \tabularnewline

\hline

\multirow{2}{*}{} & \multicolumn{1}{c|}{average} & \multicolumn{1}{c|}{manual} & \multicolumn{1}{c|}{forward} & \multicolumn{1}{c|}{success} & \multicolumn{1}{c|}{average} & \multicolumn{1}{c|}{manual} & \multicolumn{1}{c|}{forward} & \multicolumn{1}{c|}{success} & \multicolumn{1}{c|}{average} & \multicolumn{1}{c|}{manual} & \multicolumn{1}{c|}{forward} & \multicolumn{1}{c}{success} \tabularnewline
& \multicolumn{1}{c|}{return} & \multicolumn{1}{c|}{resets} & \multicolumn{1}{c|}{share} & \multicolumn{1}{c|}{rate} & \multicolumn{1}{c|}{return} & \multicolumn{1}{c|}{resets} & \multicolumn{1}{c|}{share} & \multicolumn{1}{c|}{rate} & \multicolumn{1}{c|}{return} & \multicolumn{1}{c|}{resets} & \multicolumn{1}{c|}{share} & \multicolumn{1}{c}{rate} \tabularnewline

\hline

\textbf{LNT} & \centering \textbf{33.28} & \centering \textbf{359} & \centering \textbf{75.4\%} & \centering \textbf{99.1\%} & \centering \textbf{44.57} & \centering \textbf{1714} & \centering \textbf{48.9\%} & \centering \textbf{93.1\%} & \centering \textbf{6.65} & \centering \textbf{1195} & \centering \textbf{5.7\%} & \centering \textbf{95.7\%} \tabularnewline
\textbf{ours} & \centering 17.75 & \centering 1231 & \centering 47.9\% & \centering 94.9\% & \centering 30.27 & \centering 2525 & \centering 32.6\% & \centering 71.2\% &  \centering 6.29 & \centering 1305 & \centering 4.2\% & \centering 68.7\% \tabularnewline
\textbf{LNT-sparse} & \centering 31.70 & \centering 1527 & \centering 52.5\% & \centering 87.8\% & \centering 0.46 & \centering 7407 & \centering 0.6\% & \centering 35.3\% &  \centering 0.00 & \centering 1415 & \centering 0.1\% & \centering 94.9\% \tabularnewline

\bottomrule

\end{tabularx}
\end{center}
\vspace{-0.6cm}
\end{table*}

We compare our method against LNT introduced in prior work \cite{iros2020:eysenbach2018leave}. We set the oracle (LNT) as the original algorithm with full access to handcrafted reset rewards and set the baseline (LNT-sparse) as LNT but with access to sparse (0 or 1) reset rewards. LNT represents a best-case scenario where the reset agent receives a rich learning signal from the environment. The reset reward for the oracle was shaped in a similar fashion to the forward reward. LNT-sparse is under the same constraint as our method and is only notified when it has successfully reset. However, unlike our method, LNT-sparse does not have the benefit of example-based resets and should represent the worst-case scenario. For LNT and LNT-sparse, ensemble minimum was used to trigger resets and $Q_{thresh}$ was set to 20 and 0.1 for LNT and LNT-sparse, respectively. Ensemble size of $K=20$ was used for the reset Q-function of LNT and LNT-sparse following the default settings in \cite{iros2020:eysenbach2018leave}, giving a slight advantage over our method which used $K=5$ for the reset success classifier.

\newcommand{\betathreshsubpath}{eval}

\begin{figure}[b]
    \vspace{\reducemargin}
    \centering
    \begin{subfigure}[b]{0.493\linewidth}
        \centering
        \includegraphics[width = 1\linewidth]{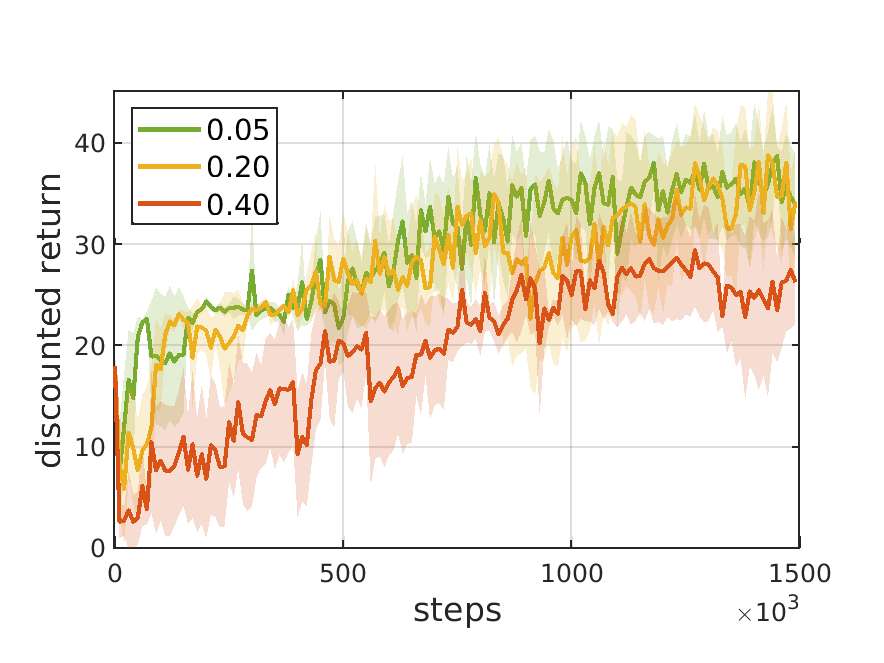}
    \end{subfigure}
    \hfill
    \begin{subfigure}[b]{0.493\linewidth}
        \centering
        \includegraphics[width = 1\linewidth]{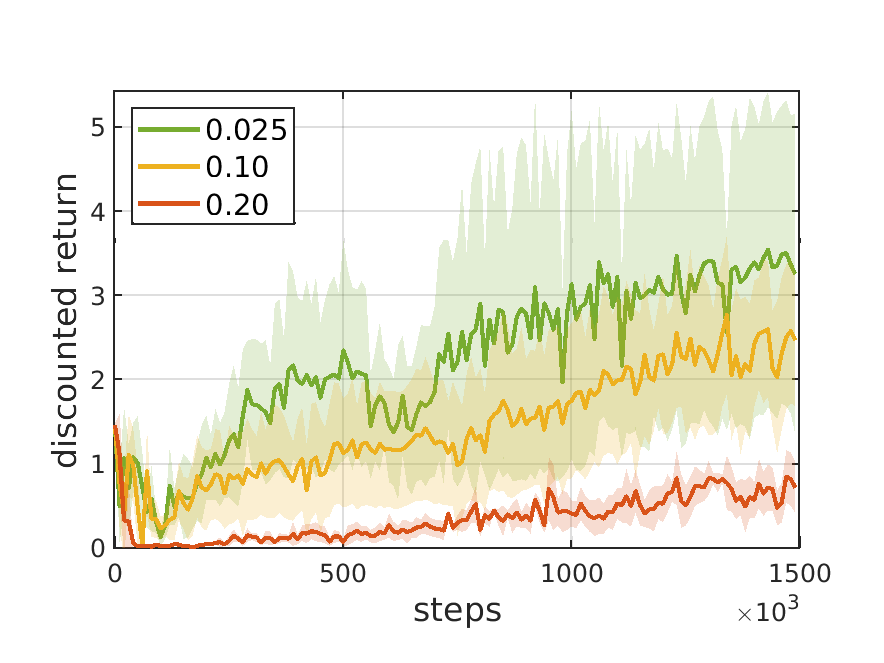}
    \end{subfigure}
    \begin{subfigure}[b]{0.493\linewidth}
        \centering
        \includegraphics[width = 1\linewidth]{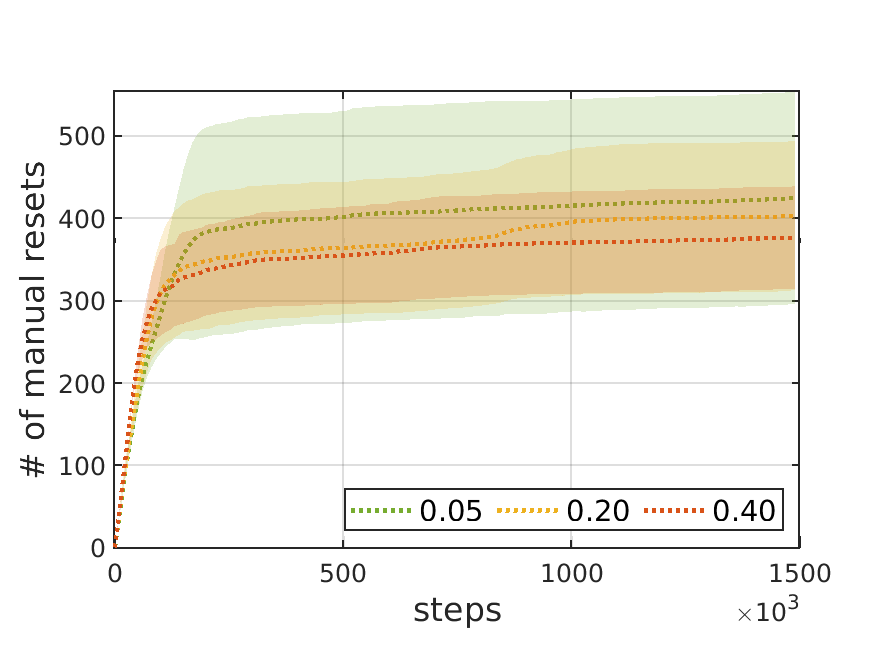}
        \caption{\textbf{\textit{peg-insertion (insert)}}}
        \label{fig:beta_thresh(peg-insertion_remove)}
    \end{subfigure}
    \hfill
    \begin{subfigure}[b]{0.493\linewidth}
        \centering
        \includegraphics[width = 1\linewidth]{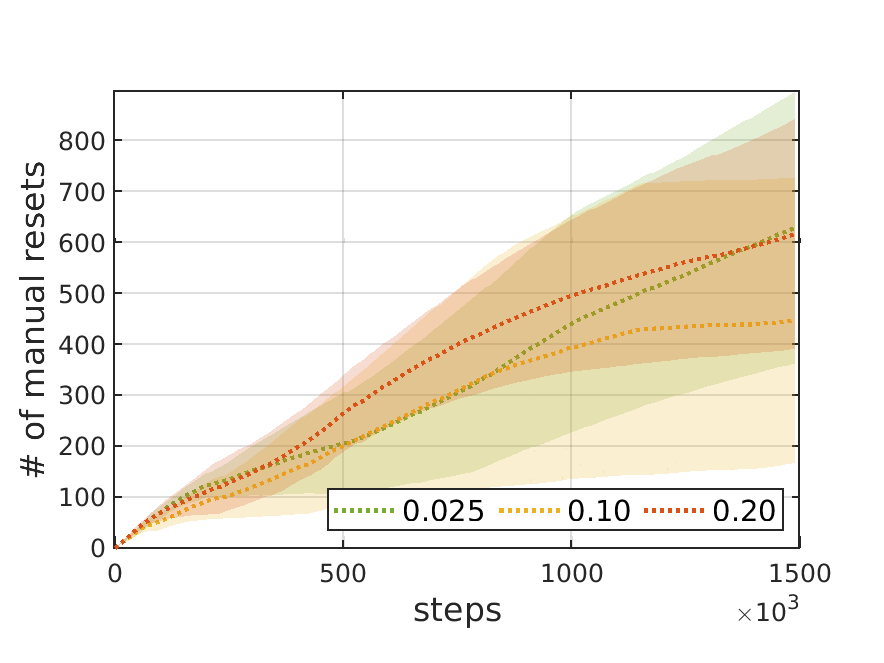}
        \caption{\textbf{\textit{cliff-cheetah (move forward)}}}
        \label{fig:beta_thresh(cliff-cheetah)}
    \end{subfigure}
    \caption{Effect of $p_{thresh}$ on the learning curve.}
    \label{fig:beta_thresh}
    \vspace{\reducemargin}
\end{figure}

Overall, our method outperformed LNT-sparse and in some instances performed as well or better than LNT, in terms of the average forward episodic return (higher is better) and the number of manual resets (lower is better) as shown in Table \ref{table:results}. We compare the average of the episodic return throughout training instead of the final episodic return to consider the convergence rate. \textbf{\textit{ball-in-cup}} is the easiest environment and LNT-sparse performed as well or better than our method in terms of the average episodic return albeit with a slower convergence rate and with more manual resets. For \textbf{\textit{peg-insertion}}, which is a manipulation environment with greater difficulty, our method vastly outperformed LNT-sparse and performed close to LNT. Similar can be said for the cliff environments where our method mostly outperformed the LNT-sparse and performed on par with LNT. Our method even outperformed LNT for some tasks which may be attributed to either the relative ease of the task or the difficulty of shaping the reset reward.

Considering the ratio of steps taken by the forward agent versus total steps taken (forward share) and the share of successful reset attempts by the reset agent (success rate) provides additional insight. A common failure mode is when the reset agent becomes overly pessimistic and prematurely triggers a reset, preventing the forward agent from improving. In this case, the success rate goes up since the resets become very easy but the forward share suffers. This is indeed the case for LNT-sparse, where the forward share is in the single digits ($\sim$2\%) while the success rate remains somewhat comparable to our method and LNT (Table \ref{table:results}) for the \textbf{\textit{peg-insertion}} and \textbf{\textit{cliff-walker}} environments. The low forward share indicates a near-immediate interruption of the forward agent which explains the exceptionally low average episodic return of LNT-sparse.

\newcommand{\distsubpath}{train}

\begin{figure}[b]
    \vspace{\reducemargin}
    \centering
    \begin{subfigure}[b]{0.493\linewidth}
        \centering
        \includegraphics[width = 1\linewidth]{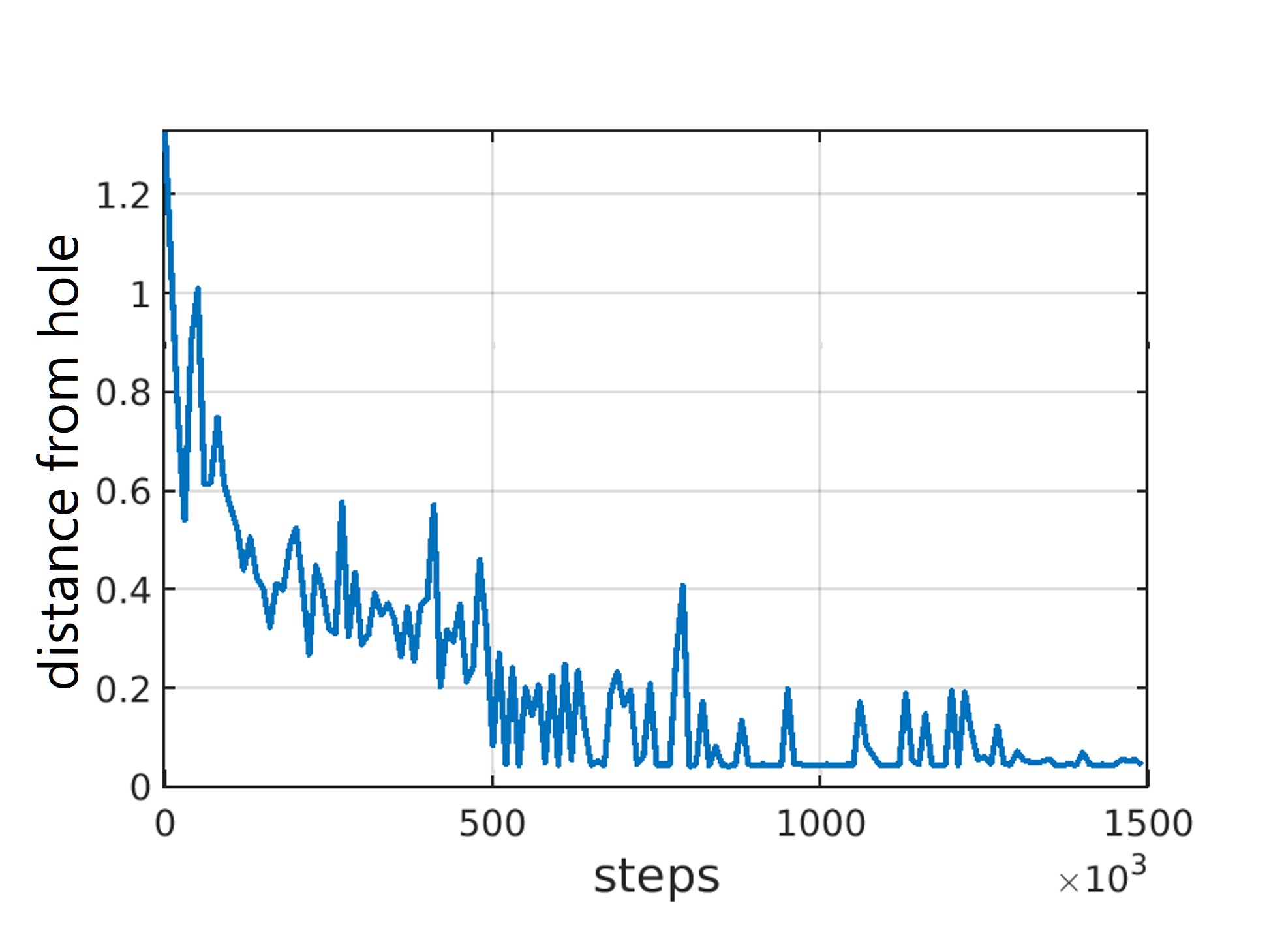}
        \caption{\textbf{\textit{peg-insertion (insert)}}}
        \label{fig:dist(ball-in-cup)}
    \end{subfigure}
    \begin{subfigure}[b]{0.493\linewidth}
        \centering
        \includegraphics[width = 1\linewidth]{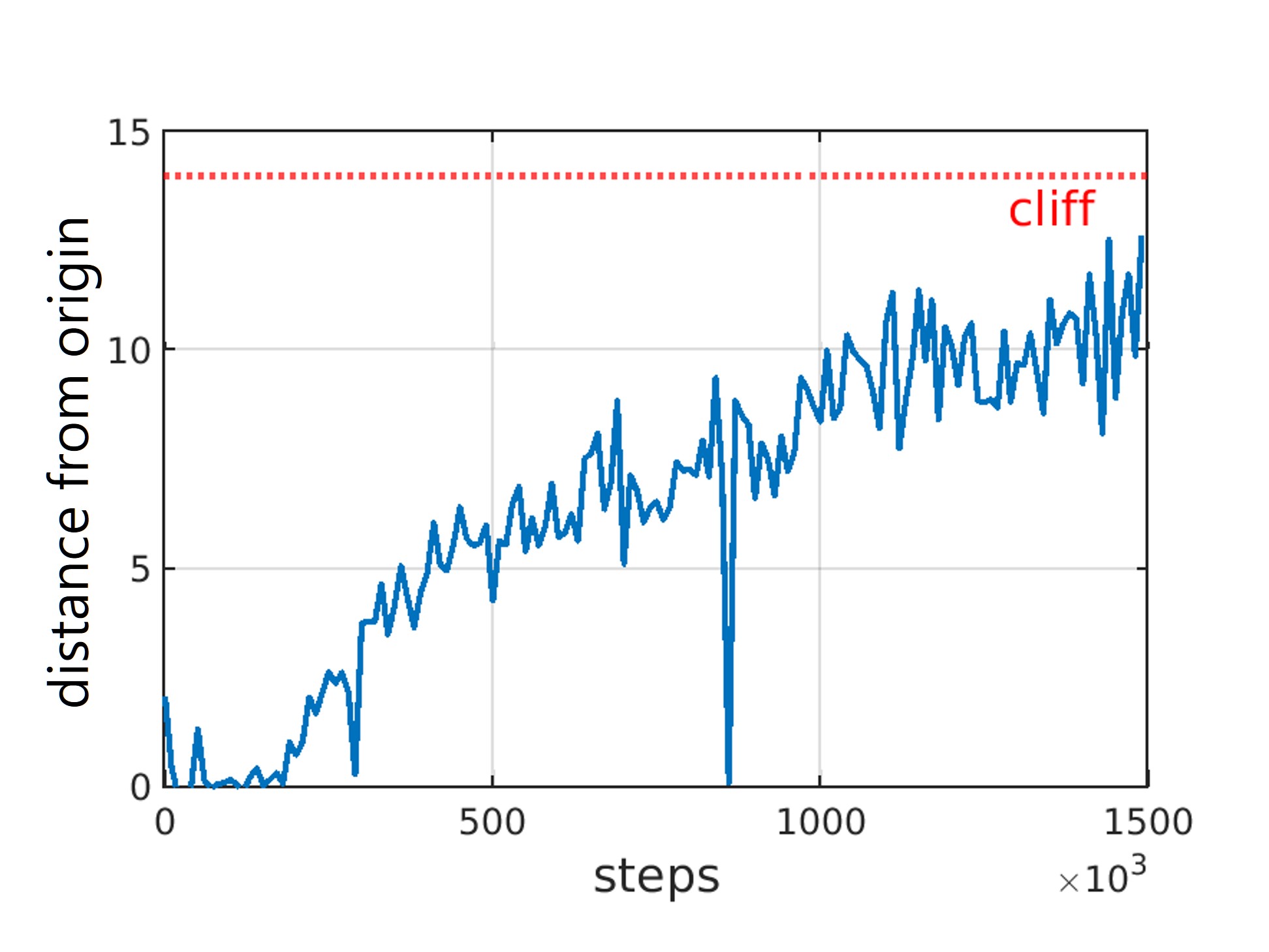}
        \caption{\textbf{\textit{cliff-cheetah}}}
        \label{fig:dist(peg-insertion)}
    \end{subfigure}
    \caption{Curriculum for the forward agent with reset trigger.}
    \label{fig:dist}
    \vspace{\reducemargin}
\end{figure}

\subsubsection{Adjusting the threshold for reset trigger}

Resets are triggered if the ensemble average of the probability of successful reset given current state and action is below some threshold $p_{thresh}$. Lower values of $p_{thresh}$ indicate a greater ``step budget'' for the reset agent and vice versa. Thus, a smaller $p_{thresh}$ allows the forward agent to venture further while a larger $p_{thresh}$ confines the forward agent to states near the initial states. Varying $p_{thresh}$ reveals such a trend in the learning curves shown in Fig. \ref{fig:beta_thresh}. As $p_{thresh}$ increases, forward episodic return is stunted for both environments but more so for \textbf{\textit{cliff-cheetah}}, since $p_{thresh}$ effectively limits the distance travelled along the $x$ axis. In terms of the number of manual resets, higher values of $p_{thresh}$ prevent more manual resets in the case of \textbf{\textit{peg-insertion (insert)}} albeit the differences are minor. For \textbf{\textit{cliff-cheetah}}, the optimal threshold that fully prevents manual resets exist at $p_{thresh}=0.10$, and values higher or lower result in worse performance. One likely explanation is that lower values do not adequately prevent the forward agent from falling off the cliff and higher values trigger frequent resets leading to higher counts of manual resets even if the success rate might be similar.

\newcommand{\hardwaredemosubpath}{demo}

\begin{figure*}[t]
    \centering
    \begin{subfigure}[b]{1\linewidth}
        \centering
        \includegraphics[width = 1\linewidth]{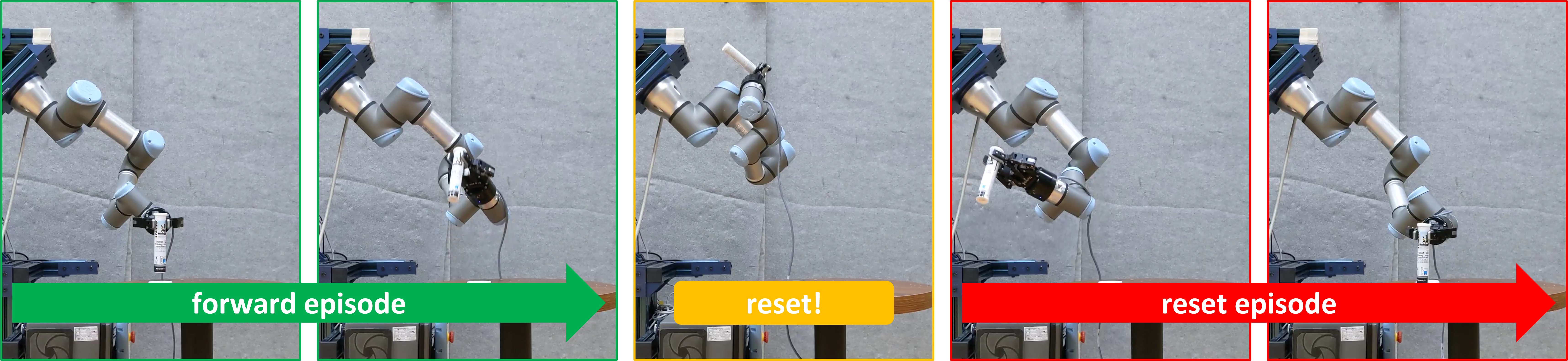}
        \caption{\textbf{\textit{ur3-peg}}}
        \label{fig:ur3-peg(demo)}
        \vspace{0.2cm}
    \end{subfigure}
    \begin{subfigure}[b]{1\linewidth}
        \centering
        \includegraphics[width = 1\linewidth]{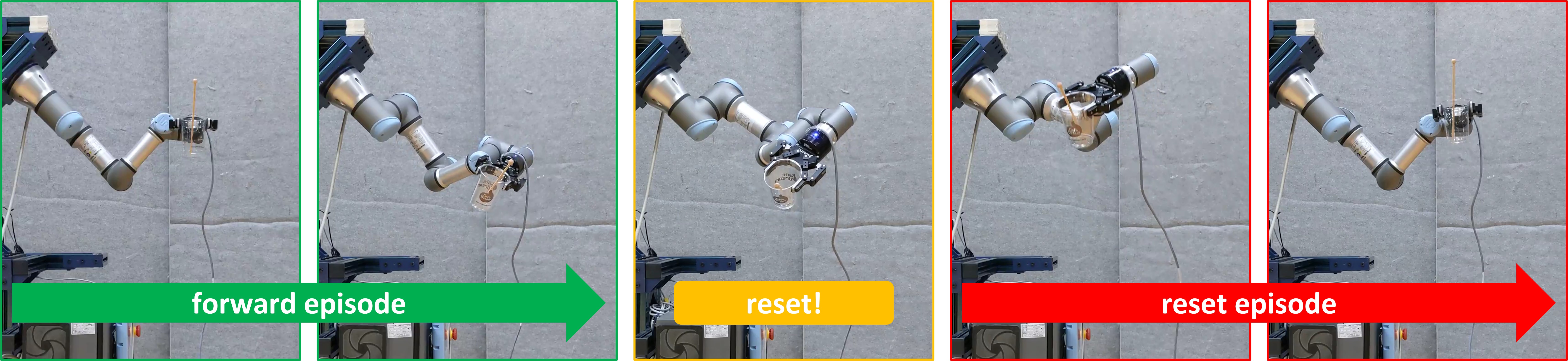}
        \caption{\textbf{\textit{ur3-reacher}}}
        \label{fig:ur3-reacher(demo)}
    \end{subfigure}
    \caption{Forward and reset episodes from real robot training (see supplementary video). Forward and reset agents were learned from scratch on hardware with a training time of 160 min and 80 min for \textbf{\textit{ur3-peg}} and \textbf{\textit{ur3-reacher}}, respectively.}
    \label{fig:ur3(demo)}
    \vspace{\reducemargin}
\end{figure*}

\subsubsection{Implicit curriculum for the forward agent}

The reset trigger mechanism enforces an implicit curriculum for the forward agent, acting as a safety layer to prevent irrecoverable states. Fig. \ref{fig:dist} demonstrates how a curriculum is imposed on the forward agent by plotting the state of the environment right before a reset trigger. In the early stages of training, \textbf{\textit{peg-insertion(insert)}} is prevented from straying far from the initial position resulting in a large peg-to-hole distance, but as training progresses it is allowed to move further until it eventually succeeds in inserting the peg. Similarly, \textbf{\textit{cliff-cheetah}} moves further away from the origin as training progresses but only up to a point right before the cliff as the reset agent realizes that it is physically impossible to return to the initial state after falling off the cliff. Without the reset agent, \textbf{\textit{cliff-cheetah}} would blindly maximize its forward objective of moving forward with no regard for the consequence.

\subsection{Real robot tasks}

To demonstrate the potential of our method to real robot learning, we learn from scratch on a 6DoF UR3 manipulator. The 12-dimensional state vector of the UR3 environment was defined as the cosines and sines of the joint angles to prevent discontinuity and bound the values to $[-1,1]$. Since URScript API provides limited control modes, a 6-dimensional action vector was defined to be compatible with the \textit{speedj} API, which takes the desired joint velocity as the input. The manipulator was mounted downwards at a 45-degree angle to mimic the configuration of a human arm and was operated at 25 Hz. Two tasks were designed for the UR3 environment. Inspired by \textbf{\textit{peg-insertion(remove)}}, \textbf{\textit{ur3-peg}} is tasked with placing the 30 cm peg attached to its end-effector. \textbf{\textit{ur3-reacher}} is tasked with reaching a goal 40 cm away from the initial position but with one caveat. It must do so while holding a cup filled with a top-heavy stick such that spilling the stick would result in an irrecoverable state. Thus, \textbf{\textit{ur3-reacher}} has an additional state dimension that encodes the status of the stick. Note that the forward reward of \textbf{\textit{ur3-reacher}} only considers the position (not orientation) of the end-effector such that the burden of preventing irrecoverable states lies on the reset agent.

\newcommand{\hardwaresubpath}{train_blueblack}

\begin{figure}[b]
    \vspace{\reducemargin}
    \centering
    \begin{subfigure}[b]{0.493\linewidth}
        \centering
        \includegraphics[width = 1\linewidth]{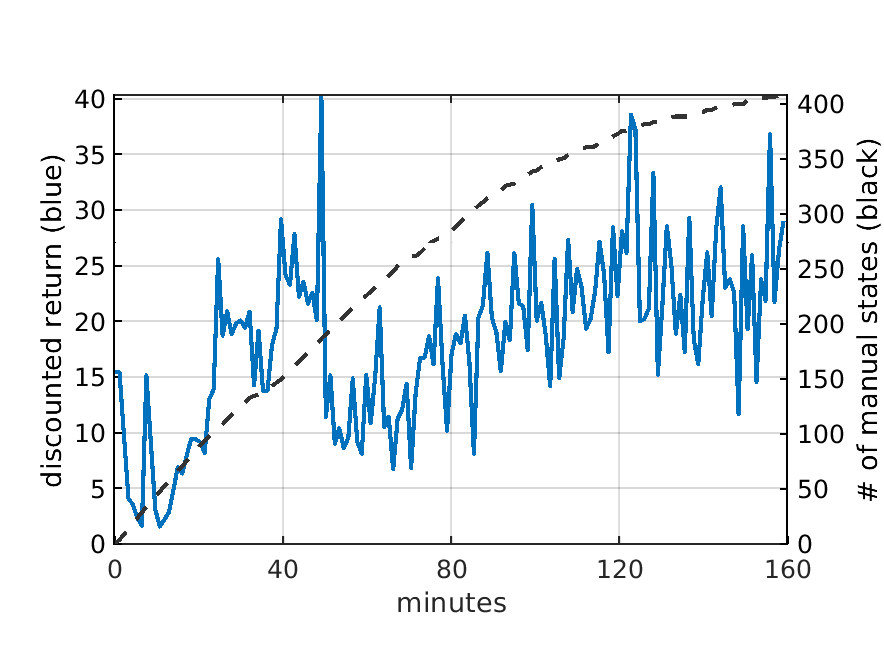}
        \caption{\textbf{\textit{ur3-peg}}}
        \label{fig:ur3-peg(training)}
    \end{subfigure}
    \begin{subfigure}[b]{0.493\linewidth}
        \centering
        \includegraphics[width = 1\linewidth]{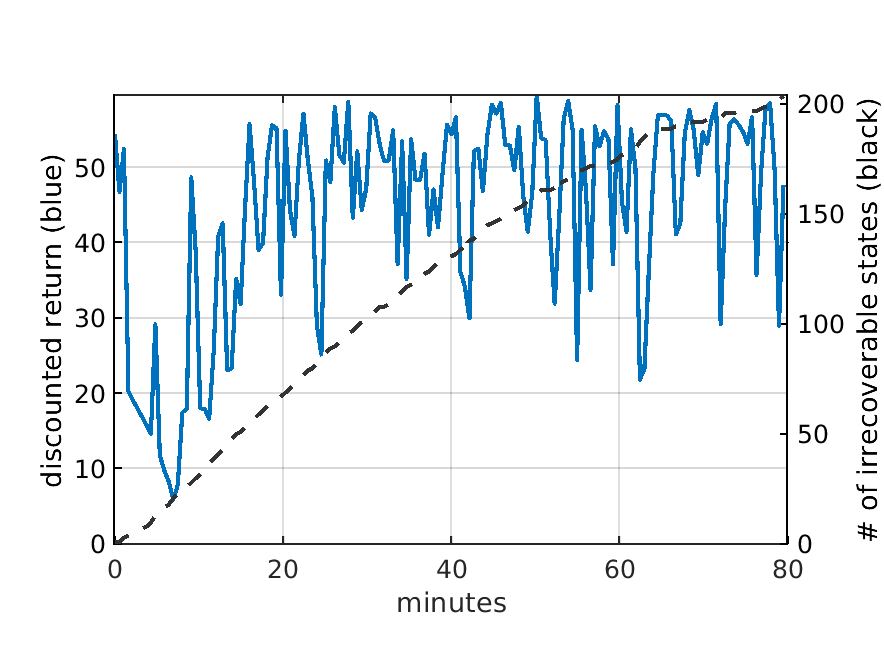}
        \caption{\textbf{\textit{ur3-reacher}}}
        \label{fig:ur3-reacher(training)}
    \end{subfigure}
    \caption{Real-time training curve for the UR3 environments.}
    \label{fig:ur3(training)}
    \vspace{\reducemargin}
\end{figure}

\subsection{Real robot results}

We applied our method to two UR3 environments and showcase snapshots of a pair of forward and reset episodes from training (Fig. \ref{fig:ur3(demo)}). The reset agent successfully learns to place the peg back to the origin for \textbf{\textit{ur3-peg}} and learns to return the cup to the initial position without spilling the stick for \textbf{\textit{ur3-reacher}}. We recommend the reader to check out the supplementary video. We also plot the forward episodic return for both environments and plot the number of manual resets or the number of irrecoverable states (Fig. \ref{fig:ur3(training)}). We specifically plot the number of irrecoverable states for \textbf{\textit{ur3-reacher}}, since irrecoverable states will likely require human intervention which severely hinders autonomy compared to other modes of reset failures.

The reset agent learns to successfully reset for both environments (up to 90\% success rate) with the number of manual resets or the number of irrecoverable states plateauing after a training time of 160 minutes (1718 episodes) and 80 minutes (644 episodes) for \textbf{\textit{ur3-peg}} and \textbf{\textit{ur3-reacher}}, respectively. Corresponding wall times, which include time for manual resets, were 188 minutes and 95 minutes, achieving a train time efficiency of 85\% and 84\% made possible by learning during both forward and reset episodes. Note that the reset agent starts with zero examples of initial states for the sake of autonomous learning and collecting initial states before training could yield even better results. In terms of the forward episodic return, both environments display a similar trend. There is a dip in return when the reset agent frequently interrupts the forward agent to prevent un-resettable states, followed by a gradual increase as the forward agent learns to achieve the forward task without triggering resets.

\section{Conclusion}
We proposed a self-supervised reset agent learning framework based on example-based control to impose an implicit curriculum and provide a safety layer for the conventional RL agent in an attempt towards fully autonomous learning agents. To that end, we trained a reset success classifier to learn how to reset and trigger resets based on the probability of successful reset to decide when to reset. We applied our method to various simulated tasks to demonstrate that our method allows the forward agent to improve gradually while reducing the number of manual resets. We confirmed that our method vastly outperforms sparse-LNT (baseline) and performs comparably to LNT (oracle) over several performance metrics. We also discussed the effects of $p_{thresh}$ and how reset triggers provide a curriculum for the forward agent. Finally, we deployed our method on a UR3 manipulator to demonstrate the potential of the method for autonomous robot learning in the real world.

The success of our method hinges on striking the right balance between the two competing objectives---improving the forward agent and guaranteeing successful resets---which was critical for the more difficult tasks. We hope to build upon our work in the future by eliminating the need for notifying the reset agent when it has reset by learning a dynamics-aware distance metric between states. We also plan to expand our work to the multi-task setting where a single reset agent is coupled with multiple forward agents or a goal-conditioned forward agent.










\bibliographystyle{./bibtex/IEEEtran}
\bibliography{./bibtex/myIEEE, ./bibtex/mybib}

\begin{thebibliography}{10}
\providecommand{\url}[1]{#1}
\csname url@rmstyle\endcsname
\providecommand{\newblock}{\relax}
\providecommand{\bibinfo}[2]{#2}
\providecommand\BIBentrySTDinterwordspacing{\spaceskip=0pt\relax}
\providecommand\BIBentryALTinterwordstretchfactor{4}
\providecommand\BIBentryALTinterwordspacing{\spaceskip=\fontdimen2\font plus
\BIBentryALTinterwordstretchfactor\fontdimen3\font minus
  \fontdimen4\font\relax}
\providecommand\BIBforeignlanguage[2]{{%
\expandafter\ifx\csname l@#1\endcsname\relax
\typeout{** WARNING: IEEEtran.bst: No hyphenation pattern has been}%
\typeout{** loaded for the language `#1'. Using the pattern for}%
\typeout{** the default language instead.}%
\else
\language=\csname l@#1\endcsname
\fi
#2}}

\bibitem{iros2020:duan2016benchmarking}
Y.~Duan, \emph{et~al.}, ``Benchmarking deep reinforcement learning for
  continuous control,'' in \emph{International Conference on Machine Learning},
  2016, pp. 1329--1338.

\bibitem{iros2020:gu2017deep}
S.~Gu, \emph{et~al.}, ``Deep reinforcement learning for robotic manipulation
  with asynchronous off-policy updates,'' in \emph{2017 IEEE international
  conference on robotics and automation (ICRA)}.\hskip 1em plus 0.5em minus
  0.4em\relax IEEE, 2017, pp. 3389--3396.

\bibitem{iros2020:vinyals2019grandmaster}
O.~Vinyals, \emph{et~al.}, ``Grandmaster level in starcraft ii using
  multi-agent reinforcement learning,'' \emph{Nature}, vol. 575, no. 7782, pp.
  350--354, 2019.

\bibitem{iros2020:berner2019dota}
C.~Berner, \emph{et~al.}, ``Dota 2 with large scale deep reinforcement
  learning,'' \emph{arXiv preprint arXiv:1912.06680}, 2019.

\bibitem{iros2020:andrychowicz2020learning}
O.~M. Andrychowicz, \emph{et~al.}, ``Learning dexterous in-hand manipulation,''
  \emph{The International Journal of Robotics Research}, vol.~39, no.~1, pp.
  3--20, 2020.

\bibitem{ral2021:yahya2017collective}
A.~Yahya, \emph{et~al.}, ``Collective robot reinforcement learning with
  distributed asynchronous guided policy search,'' in \emph{2017 IEEE/RSJ
  International Conference on Intelligent Robots and Systems (IROS)}.\hskip 1em
  plus 0.5em minus 0.4em\relax IEEE, 2017, pp. 79--86.

\bibitem{ral2021:sharma2020emergent}
A.~Sharma, \emph{et~al.}, ``Emergent real-world robotic skills via unsupervised
  off-policy reinforcement learning,'' \emph{arXiv preprint arXiv:2004.12974},
  2020.

\bibitem{ral2021:kalashnikov2018qt}
D.~Kalashnikov, \emph{et~al.}, ``Qt-opt: Scalable deep reinforcement learning
  for vision-based robotic manipulation,'' \emph{arXiv preprint
  arXiv:1806.10293}, 2018.

\bibitem{iros2020:zeng2019tossingbot}
A.~Zeng, \emph{et~al.}, ``Tossingbot: Learning to throw arbitrary objects with
  residual physics,'' in \emph{Proceedings of Robotics: Science and Systems},
  2019.

\bibitem{ral2021:ha2021learning}
S.~Ha, \emph{et~al.}, ``Learning to walk in the real world with minimal human
  effort,'' in \emph{Conference on Robot Learning}.\hskip 1em plus 0.5em minus
  0.4em\relax PMLR, 2021, pp. 1110--1120.

\bibitem{ral2021:xu2020continual}
K.~Xu, \emph{et~al.}, ``Continual learning of control primitives: Skill
  discovery via reset-games,'' \emph{Advances in Neural Information Processing
  Systems}, vol.~33, pp. 4999--5010, 2020.

\bibitem{ral2021:zhu2020ingredients}
\BIBentryALTinterwordspacing
H.~Zhu, \emph{et~al.}, ``The ingredients of real world robotic reinforcement
  learning,'' in \emph{International Conference on Learning Representations},
  2020. [Online]. Available: \url{https://openreview.net/forum?id=rJe2syrtvS}
\BIBentrySTDinterwordspacing

\bibitem{ral2021:gupta2021reset}
A.~Gupta, \emph{et~al.}, ``Reset-free reinforcement learning via multi-task
  learning: Learning dexterous manipulation behaviors without human
  intervention,'' in \emph{2021 IEEE International Conference on Robotics and
  Automation (ICRA)}.\hskip 1em plus 0.5em minus 0.4em\relax IEEE, 2021, pp.
  6664--6671.

\bibitem{ral2021:sharma2021persistent}
A.~Sharma, \emph{et~al.}, ``Autonomous reinforcement learning via subgoal
  curricula,'' in \emph{Advances in Neural Information Processing Systems},
  2021.

\bibitem{ral2021:eysenbach2020rewriting}
B.~Eysenbach, \emph{et~al.}, ``Rewriting history with inverse rl: Hindsight
  inference for policy improvement,'' \emph{Advances in neural information
  processing systems}, vol.~33, pp. 14\,783--14\,795, 2020.

\bibitem{ral2021:han2015learning}
W.~Han, S.~Levine, and P.~Abbeel, ``Learning compound multi-step controllers
  under unknown dynamics,'' in \emph{2015 IEEE/RSJ International Conference on
  Intelligent Robots and Systems (IROS)}.\hskip 1em plus 0.5em minus
  0.4em\relax IEEE, 2015, pp. 6435--6442.

\bibitem{iros2020:eysenbach2018leave}
B.~Eysenbach, \emph{et~al.}, ``Leave no trace: Learning to reset for safe and
  autonomous reinforcement learning,'' in \emph{International Conference on
  Learning Representations (ICLR 2018)}, 2018.

\bibitem{ral2021:smith2020avid}
L.~Smith, \emph{et~al.}, ``{Avid:} learning multi-stage tasks via pixel-level
  translation of human videos,'' in \emph{Robotics: Science and Systems XVI,
  Virtual Event / Corvalis, Oregon, USA, July 12-16, 2020}, 2020.

\bibitem{ral2021:thananjeyan2021recovery}
B.~Thananjeyan, \emph{et~al.}, ``Recovery rl: Safe reinforcement learning with
  learned recovery zones,'' \emph{IEEE Robotics and Automation Letters},
  vol.~6, no.~3, pp. 4915--4922, 2021.

\bibitem{ral2021:schaul2016prioritized}
T.~Schaul, \emph{et~al.}, ``Prioritized experience replay,'' in \emph{4th
  International Conference on Learning Representations, {ICLR} 2016, San Juan,
  Puerto Rico, May 2-4, 2016, Conference Track Proceedings}, 2016.

\bibitem{ral2021:rahaman2020learning}
N.~Rahaman, \emph{et~al.}, ``Learning the arrow of time for problems in
  reinforcement learning,'' in \emph{International Conference on Learning
  Representations}, 2020.

\bibitem{ral2021:grinsztajn2021there}
N.~Grinsztajn, \emph{et~al.}, ``There is no turning back: A self-supervised
  approach for reversibility-aware reinforcement learning,'' \emph{Advances in
  Neural Information Processing Systems}, vol.~34, 2021.

\bibitem{iros2020:florensa2017reverse}
C.~Florensa, \emph{et~al.}, ``Reverse curriculum generation for reinforcement
  learning,'' in \emph{Proceedings of the 1st Annual Conference on Robot
  Learning}, ser. Proceedings of Machine Learning Research, vol.~78.\hskip 1em
  plus 0.5em minus 0.4em\relax PMLR, 13--15 Nov 2017, pp. 482--495.

\bibitem{iros2020:zakka2019form2fit}
K.~Zakka, \emph{et~al.}, ``Form2fit: Learning shape priors for generalizable
  assembly from disassembly,'' in \emph{2020 IEEE International Conference on
  Robotics and Automation (ICRA)}.\hskip 1em plus 0.5em minus 0.4em\relax IEEE,
  2020.

\bibitem{iros2020:fu2018variational}
J.~Fu, \emph{et~al.}, ``Variational inverse control with events: A general
  framework for data-driven reward definition,'' in \emph{Advances in Neural
  Information Processing Systems}, 2018, pp. 8538--8547.

\bibitem{ral2021:eysenbach2021replacing}
B.~Eysenbach, S.~Levine, and R.~R. Salakhutdinov, ``Replacing rewards with
  examples: Example-based policy search via recursive classification,''
  \emph{Advances in Neural Information Processing Systems}, vol.~34, 2021.

\bibitem{iros2020:lillicrap2015continuous}
T.~P. Lillicrap, \emph{et~al.}, ``Continuous control with deep reinforcement
  learning,'' \emph{4th International Conference on Learning Representations,
  {ICLR} 2016, San Juan, Puerto Rico, May 2-4, 2016}, 2016.

\bibitem{iros2020:osband2018randomized}
I.~Osband, J.~Aslanides, and A.~Cassirer, ``Randomized prior functions for deep
  reinforcement learning,'' in \emph{Advances in Neural Information Processing
  Systems}, 2018, pp. 8617--8629.

\bibitem{iros2020:openaigym2016}
G.~Brockman, \emph{et~al.}, ``Openai gym,'' 2016.

\bibitem{iros2020:dmcontrol2018}
Y.~Tassa, \emph{et~al.}, ``Deep{Mind} control suite,''
  https://arxiv.org/abs/1801.00690, DeepMind, Tech. Rep., 2018.

\end{thebibliography}

\end{document}